\documentclass[11pt]{article}

\usepackage[preprint]{acl}

\usepackage{times}
\usepackage{latexsym}
\usepackage[T1]{fontenc}
\usepackage[utf8]{inputenc}
\usepackage{microtype}
\usepackage{inconsolata}

\usepackage{graphicx}
\usepackage{subcaption}
\usepackage{booktabs}
\usepackage{multirow}

\usepackage{algorithm}
\usepackage{algorithmic}

\usepackage{amsmath}
\usepackage{amssymb}
\usepackage{mathtools}
\usepackage{amsthm}

\usepackage{enumitem}
\usepackage[capitalize,noabbrev]{cleveref}
\usepackage[disable,textsize=tiny]{todonotes}
\usepackage{placeins}    

\theoremstyle{plain}

\theoremstyle{definition}

\theoremstyle{remark}

\title{Dropping the Anchor: Statistical Context Summarization for
  Distributed Systems via Pulsar Attention}

\author{
  \textbf{Aryan Sood} \\
  Indian Institute of Technology Roorkee, India \\
  \texttt{aryan\_s2@ee.iitr.ac.in} \\
  \And
  \textbf{Shantanu Acharya} \\
  NVIDIA \\
  \texttt{shantanua@nvidia.com} \\
  \AND
  \textbf{Gaurav Kumar Nayak} \\
  Indian Institute of Technology Roorkee, India \\
  \texttt{gauravkumar.nayak@mfs.iitr.ac.in} \\
}

\begin{document}
\maketitle

\begin{abstract}
Inference with large language models (LLMs) on long sequences is computationally expensive due to the quadratic complexity of self-attention. Distributed blockwise methods such as Star Attention reduce this cost by sharding context across hosts, but rely on prepending a static, content-blind copy of the first block to every host. We propose Pulsar Attention, which replaces the static anchor with two lightweight, content-aware components: a small attention-sink prefix that stabilizes softmax, and compact cross-block summaries built via a Max-IDF heuristic that selects chunks containing globally rare tokens. This reduces the Phase 1 per-GPU FLOPs by up to $3.3\times$ over Star Attention while retaining an identical KV cache footprint. On RULER with Llama-3.1-8B-Instruct, Pulsar Attention outperforms Star Attention at sequence lengths up to 128K tokens and remains competitive with dense attention across most tasks, with task-dependent absolute gains of up to $4.7\%$ over the dense baseline.
\end{abstract}              
\begin{figure*}[t]
  \vskip 0.2in
  \centering
  \begin{subfigure}[t]{0.48\textwidth}
    \centering
    \begin{minipage}[c][5.8cm][c]{\linewidth}
      \centering
      \includegraphics[width=\linewidth,height=5.8cm,keepaspectratio]{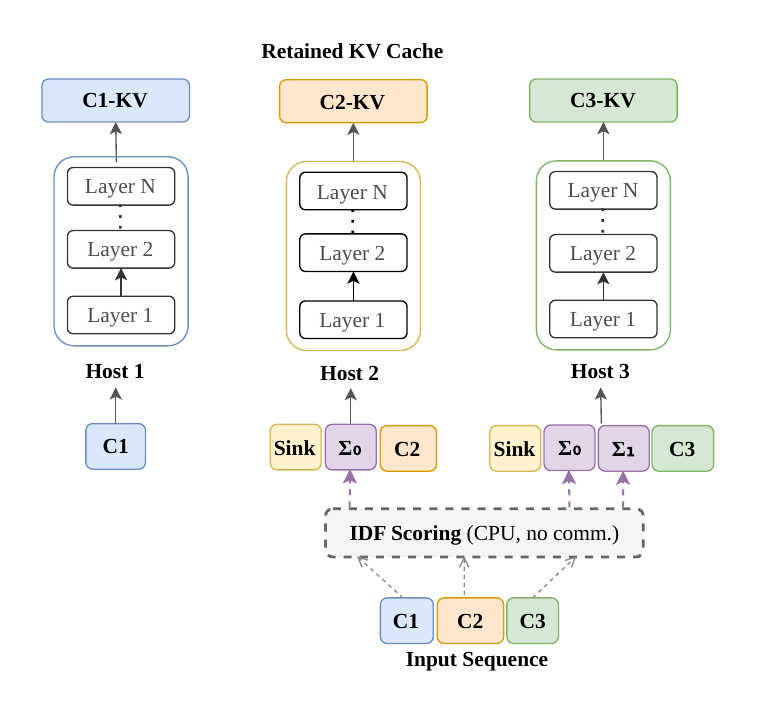}
    \end{minipage}
    \caption{Phase~1: Statistical context encoding. Host~$i$ assembles
      input $[\mathcal{S}_{\text{sink}} \parallel \Sigma_0 \parallel \cdots
      \parallel \Sigma_{i-1} \parallel B_i]$, runs a forward pass, then
      discards sink and summary KV entries, retaining only $\text{KV}(B_i)$.}
    \label{fig:phase1}
  \end{subfigure}
  \hfill
  \begin{subfigure}[t]{0.48\textwidth}
    \centering
    \begin{minipage}[c][5.8cm][c]{\linewidth}
      \centering
      \includegraphics[width=\linewidth,height=5.8cm,keepaspectratio]{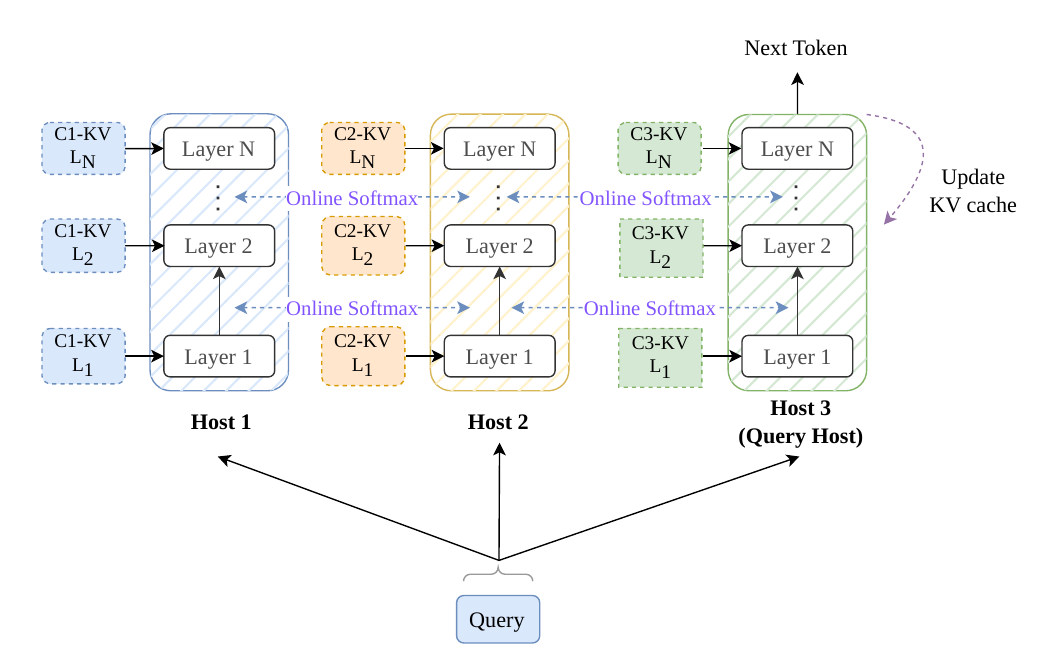}
    \end{minipage}
    \caption{Phase~2: Global query encoding. Each host attends over its
      retained KV shard; an \texttt{all\_gather} and online softmax
      merge~\citep{milakov2018online} recovers exact global attention
      without transmitting the full KV cache.}
    \label{fig:phase2}
  \end{subfigure}
  \caption{Pulsar Attention overview. \textit{(Left)}~Phase~1 replaces the
    static anchor block with a content-aware prefix: a small attention-sink
    and Max-IDF statistical summaries of all causally preceding blocks.
    \textit{(Right)}~Phase~2 broadcasts the query to all hosts and merges
    local attention scores into an exact global output via online softmax.}
  \label{fig:pulsar-overview}
  \vskip -0.1in
\end{figure*}

\section{Introduction}
\label{sec:introduction}

Large Language Models (LLMs) have demonstrated strong performance on tasks requiring long context windows, including repository-level code generation, multi-document summarisation, and long-horizon retrieval. However, scaling inference to these lengths remains expensive: self-attention is quadratic in sequence length, and the Key–Value (KV) cache that avoids redundant computation during autoregressive decoding grows linearly, quickly exhausting device memory on a single GPU.

Distributed inference methods address this bottleneck by sharding context across multiple GPUs. Ring Attention \citep{liu2023ring} computes exact global attention by circulating KV blocks in a ring topology, but pays for correctness with coordinated cross-host communication at every layer. Star Attention \citep{acharya2024star} eliminates this communication during context encoding by splitting the sequence into independent blocks processed in parallel, then recovering global attention at query time through a lightweight softmax merge. The key enabler is an anchor block: a copy of the first block prepended to every host's input. This stabilizes the blockwise attention distribution. However, the anchor duplicates the same tokens regardless of their relevance, doubling per-host sequence length during context encoding and providing no information from intermediate blocks (Section \ref{sec:background}).

We propose Pulsar Attention, which replaces the static anchor with two lightweight, content-aware mechanisms: (i) a small attention-sink prefix (64 tokens from the sequence start) that stabilizes the softmax distribution at a fraction of the anchor's cost, and (ii) statistical block summaries, constructed by scoring contiguous chunks with a Max-IDF heuristic and retaining only the top-$k$ most informative chunks per block, which propagate causally from earlier blocks to later ones. Unlike the anchor, these summaries adapt to each block's content, prioritizing chunks containing tokens most likely to carry task-critical information such as entity names, identifiers, or numerical values.

The result is a strict reduction in Phase 1 compute: each host processes its local block plus a small, fixed-size prefix rather than a full duplicate block, reducing per-GPU FLOPs by up to 3.3$\times$ relative to Star Attention while retaining an identical KV cache footprint. On RULER, Pulsar Attention outperforms both Star Attention and dense attention from 32K to 128K tokens, with absolute accuracy gains of up to 4.7\% over the dense baseline, suggesting that content-aware summarization not only saves compute but actively improves attention quality by filtering out dilutive distractor tokens.

Our main contributions are:
\begin{itemize}
\item We replace Star Attention's static anchor with content-aware summaries and a small attention sink, reducing Phase 1 attention FLOPs by up to $3.3\times$ while retaining an identical KV cache footprint.
\item We introduce a Max-IDF scoring mechanism and validate it against three alternative heuristics on \textsc{BABILong} across multiple context lengths, showing that rare-token-based selection effectively captures cross-block context.
\item We demonstrate empirically on RULER that Pulsar Attention outperforms both Star Attention and dense attention at context lengths up to 128K, with gains of up to $+4.7\%$ over the dense baseline, and demonstrate competitive accuracy on Qwen3-4B-Instruct without architecture-specific tuning, confirming cross-architecture applicability.
\item We validate FLOPs-based efficiency estimates with end-to-end wall-clock profiling and GPU memory measurements, confirming speedups of up to 6.24$\times$ over dense attention at 128K.
\end{itemize}

\section{Background}
\label{sec:background}

In standard Transformer-based LLMs, autoregressive decoding relies on self-attention computed as:
\begin{equation}
  \text{Attention}(Q, K, V)
    = \text{softmax}\!\left(\frac{QK^\top}{\sqrt{d}}\right)V
\end{equation}
where $Q, K, V \in \mathbb{R}^{L \times d}$ are the query, key, and value matrices for a sequence of length $L$ and head dimension $d$. To avoid redundant recomputation, LLMs cache the key and value vectors of all previous tokens during generation. While this optimizes compute, the KV cache grows linearly with sequence length and the attention computation remains quadratic in $L$, quickly exhausting device memory and wall-clock budget at long contexts.

\citet{liu2023ring} addresses this by distributing the KV cache across GPUs and circulating blocks in a ring to compute exact global attention. This enables arbitrarily long sequences but requires cross-host communication at every layer, introducing synchronization overhead that scales with model depth.

\citet{acharya2024star} eliminates this communication during context encoding with a two-phase design. In Phase~1, the context is split into contiguous blocks processed independently on parallel hosts. Without any shared context, blockwise attention produces spurious per-block attention sinks; tokens at the start of each block accumulate disproportionate softmax mass simply by virtue of their local position, creating a degenerate attention distribution that fails to approximate global attention. To prevent this, a copy of the first block (the anchor block) is prepended to each host's input. The anchor serves a dual role: its initial tokens absorb the attention sink effect \citep{xiao2023efficient}, redirecting spurious sinks away from block-internal positions, while its semantic content provides cross-block context that later blocks need for accurate attention during Phase 2. Ablation experiments in \citet{acharya2024star} confirm that anchor content is critical; replacing it with constant or random tokens causes accuracy degradation; while the anchor's positional placement has comparatively minor effect. In Phase 2, the query is broadcast to all hosts, which compute local attention scores and aggregate them via an online softmax merge \citep{milakov2018online} to recover global attention.

While effective, the anchor is content-fixed: it always duplicates the first block regardless of its relevance to any given later block, doubling per-host sequence length and FLOPs during Phase 1. A block $i$ receives no information about intermediate blocks $2$ through $i-1$, and as sequences grow longer the gap between anchor and local block widens, meaning this static duplication captures a diminishing fraction of the cross-block context. Our method addresses this limitation.

\section{Methodology}
\label{sec:methodology}

\subsection{Phase 1: Statistical Context Encoding}
\label{sec:method-phase1}

Like Star Attention, Pulsar Attention partitions an input of length $L$ into $B$ contiguous logical blocks $\{B_0,\ldots,B_{B-1}\}$ and maps them to $H$ physical hosts through $\pi$. Logical block count and physical host count need not be equal: multiple logical blocks may be mapped to one physical host, as in the 128K configuration with $B=8$ and $H=4$ (\cref{fig:phase1}). Rather than duplicating the entire first block as a static anchor, we replace it with an attention-sink prefix and Max-IDF block summaries.

\textbf{Attention sinks.} The first $s_{\text{sink}}$ tokens of $B_0$ are extracted as a fixed prefix $\mathcal{S}_{\text{sink}}$ (default $s_{\text{sink}} = 64$). Following \citet{xiao2023efficient}, these tokens stabilize the softmax distribution during blockwise attention at a fraction of the cost of a full anchor block.

\begin{algorithm}[t]
  \caption{Pulsar Attention}
  \label{alg:pulsar}
  \begin{algorithmic}[1]
    \STATE \textbf{Input:} Logical blocks $\{B_0,\dots,B_{B-1}\}$, mapping $\pi$, query $Q$, sink size $s_{\text{sink}}$, chunk size $m$, chunks per block $k$, max tokens $T$
    \STATE \textbf{Output:} Generated token sequence

    \STATE $\mathcal{S}_{\text{sink}} \leftarrow B_0[0{:}s_{\text{sink}}]$;\; build $\mathrm{IDF}(t)$ from $\mathrm{df}(t)$ across all blocks

    \COMMENT{\textit{Phase 1: Statistical Context Encoding}}
    \FOR{logical block $i = 0,\dots,B{-}1$ (on host $\pi(i)$)}
        \STATE $\Sigma_i \leftarrow$ top-$k$ chunks of $B_i$ by
               $\max_{t\in C}\mathrm{IDF}(t)$, in positional order
        \STATE $\text{Input}_i \leftarrow [B_0]$ if $i{=}0$, else
               $[\mathcal{S}_{\text{sink}} \!\parallel\!
                \Sigma_0 \!\parallel\! \cdots \!\parallel\!
                \Sigma_{i-1} \!\parallel\! B_i]$
        \STATE $\text{KV}_i \leftarrow \textsc{ForwardPass}(\text{Input}_i)$;\;
               discard non-$B_i$ KV entries
    \ENDFOR

    \COMMENT{\textit{Phase 2: Query Encoding \& Generation}}
    \STATE Broadcast $Q$ to all hosts;\; $q \leftarrow Q$;\; $\text{output} \leftarrow [\,]$
    \FOR{$t = 1$ \textbf{to} $T$}
        \STATE $(A_h, \ell_h) \leftarrow \textsc{FlashAttn}(q, \text{KV}_h)$
               for each host $h$ in parallel;\;
               \texttt{all\_gather} $\{(A_h, \ell_h)\}$
        \STATE $(A, \ell) \leftarrow \textsc{OnlineSoftmaxMerge}
               \bigl(\{(A_h,\ell_h)\}_{h=1}^{H}\bigr)$
        \STATE $q \leftarrow \operatorname{argmax}(\textsc{LMHead}(A))$;\;
               append $\mathrm{KV}(q)$ to the designated generation shard;\;
               append $q$ to $\text{output}$;\;
               \textbf{if} $q = \texttt{EOS}$ \textbf{then break}
    \ENDFOR
    \STATE \textbf{return} output
  \end{algorithmic}
\end{algorithm}

\begin{table*}[t]
  \caption{RULER average accuracy (\%) on Llama-3.1-8B-Instruct.
    \textsuperscript{\textdagger}From \citet{acharya2024star},
    same model and setup. $B=4$ for 16K-64K; $B=8$ for 128K. \textbf{Bold} = best among efficiency-focused methods.}
  \label{tab:ruler_main}
  \begin{center}
    \begin{small}
      \begin{tabular}{lcccccc}
        \toprule
        \textbf{Method} & \textbf{Setting} &
        \textbf{16K} & \textbf{32K} & \textbf{64K} &
        \textbf{128K} & \textbf{Avg} \\
        \midrule
        Full Attention
          & Single GPU & 92.2 & 87.5 & 84.8 & 76.3 & 85.2 \\
        \midrule
        StreamingLLM\textsuperscript{\textdagger}
          & Single GPU & 74.8 & 48.6 & 26.2 & 30.8 & 45.1 \\
        MInference\textsuperscript{\textdagger}
          & Single GPU & 93.3 & 86.5 & 84.9 & 58.2 & 80.7 \\
        \midrule
        Star Attention\textsuperscript{\textdagger}
          & Multi-GPU  & 91.3 & 88.7 & 83.4 & 74.4 & 84.4 \\
        \textbf{Pulsar Attention (Ours)}
          & Multi-GPU  & 90.4 & \textbf{91.5}
                        & \textbf{89.5} & \textbf{79.1}
                        & \textbf{87.6} \\
        \bottomrule
      \end{tabular}
    \end{small}
  \end{center}
  \vskip -0.1in
\end{table*}

\textbf{Summary generation.} Before any neural computation, every host constructs an identical corpus-level IDF table from the token IDs of the full sequence, an $O(L)$ integer-only operation that requires no inter-host communication and no GPU time. Using this table, each host scores non-overlapping contiguous chunks of $m$ tokens (default $m = 32$) within each block via Max-IDF:
\begin{align}
    \text{score}(C) = \max_{t \in C} \operatorname{IDF}(t) \\[4pt]
    \operatorname{IDF}(t) = \log \left(\frac{N}{\max(\operatorname{df}(t), 1)}\right)
\end{align}

where $\operatorname{df}(t)$ is the number of blocks containing token $t$. The top-$k$ chunks per block are selected and concatenated in positional order to form block summary $\Sigma_i$, with a per-block budget of $\sigma$ tokens (default $\sigma = 0.125 \times |B_i|$). We choose contiguous chunks rather than individual tokens to preserve local syntactic structure in the resulting key–value representations. Max-IDF is preferred over averaging heuristics because identifiers, entities, and numerical values often appear as isolated high-IDF tokens that dominate retrieval relevance in long-context QA because averaging dilutes this signal by a factor of $1/m$ (Section \ref{sec:ablation-heuristic}).

\textbf{Causal Assembly.} Each host $i$ prepends only summaries from causally preceding blocks along with $\mathcal{S}_{\text{sink}}$, forming the augmented Phase~1 input:
\begin{equation}
\text{Input}_i = \begin{cases} [B_0] & i = 0 \\ [\mathcal{S}_{\text{sink}} \| \Sigma_0 \| \cdots \| \Sigma_{i-1} \| B_i] & i > 0 \end{cases}
\end{equation}
Each token retains its original global position ID, preserving the inter-token distances encoded by RoPE (see Appendix \ref{app:rope} for a comparison with contiguous re-numbering).

\textbf{KV discard.} After the forward pass, KV states for $\mathcal{S}_{\text{sink}}$ and $\{\Sigma_j\}_{j < i}$ are discarded; only local-block KV entries are retained. After discarding sink and summary KV entries, Pulsar retains only local-block KV states and therefore has the same retained distributed KV-cache footprint as Star Attention. Temporary Phase~1 activations are not claimed to be identical.

\subsection{Phase~2: Global Query Encoding and Generation}
\label{sec:method-phase2}

Because Phase~1 produces a clean, non-overlapping distributed KV-cache, Phase~2 is structurally identical to Star Attention \citep{acharya2024star}: the query is broadcast to all hosts, each computes local attention via FlashAttention, and a designated query host aggregates the results through an online-softmax merge \citep{milakov2018online}. The online-softmax merge is exact with respect to the retained distributed KV shards; Pulsar does not reproduce Dense Attention Phase~1 encoding over the original sequence. The full pseudocode is given in Algorithm~\ref{alg:pulsar}.

\section{Experiments}
\label{sec:experiments}

\subsection{Experimental Setup}

\textbf{Models.} Our primary evaluation uses Meta-Llama-3.1-8B-Instruct \citep{grattafiori2024llama}, which supports a native 128K context window. To assess cross-architecture generality, we also evaluate on Qwen3-4B-Instruct \citep{qwen3}. All experiments use bf16 precision with FlashAttention-2. Generation uses greedy decoding.

\textbf{Distributed configuration.} All distributed experiments use four physical hosts ($H=4$). At 16K--64K, $B=4$ logical blocks are mapped one per host; at 128K, $B=8$ logical blocks are mapped two per host and processed sequentially in logical-block order on each host. The summary budget is fixed at 12.5\% of block size (chunk size $m{=}32$ tokens, with $k$ scaled proportionally to context length), and the attention-sink prefix is 64 tokens. After encoding, sink and summary KV entries are discarded and only local-block KV entries are retained. The full-attention implementation and reported values were revalidated and confirmed correct.

\textbf{Benchmarks.} We evaluate on two benchmarks: (i) RULER \citep{hsieh2024ruler}, a synthetic long-context benchmark comprising 13 tasks across five categories (single-needle retrieval, multi-needle retrieval, multi-hop tracing, aggregation, and question answering), with 500 samples per task per context length; and (ii) BABILong \citep{kuratov2024babilong}, a reasoning benchmark requiring multi-step inference over supporting facts embedded in long distractor text, with 1,000 samples per task. We report exact-match accuracy for both benchmarks.

\textbf{Baselines.} We compare against four baselines: (i) Dense Attention (reported as Full Attention in tables where space requires), computed on a single GPU as the dense attention baseline; (ii) Star Attention \citep{acharya2024star}, the most directly comparable distributed method; (iii) StreamingLLM \citep{xiao2023efficient}, a sparse method combining sink tokens with sliding-window attention; and (iv) MInference \citep{jiang2024minference}, which dynamically selects per-head sparse attention patterns. Baselines marked with \textsuperscript{\textdagger} are reproduced from \citet{acharya2024star} using the same model and hardware configuration.

\subsection{Results on RULER}
\label{sec:ruler}

Table~\ref{tab:ruler_main} reports average accuracy across all 13 RULER tasks at context lengths from 16K to 128K. Pulsar Attention outperforms Star Attention at 32K (+2.8\%), 64K (+6.1\%), and 128K (+4.7\%). It also achieves higher average accuracy than Dense Attention at 32K (+4.0\%), 64K (+4.7\%), and 128K (+2.8\%). However, this average advantage is not uniform across tasks and is driven substantially by the aggregation category, particularly Common Words Extraction.

We hypothesize that summaries may reduce interference from irrelevant tokens in these aggregation settings; the current experiments do not causally establish this mechanism.

Among the single-GPU baselines, MInference performs competitively at 16K-64K but degrades sharply to 58.2\% at 128K (-20.9\% versus Pulsar), suggesting that its offline sparse pattern estimation becomes unreliable at extreme lengths. StreamingLLM collapses past 16K, confirming that recency-biased context windows cannot substitute for cross-block information flow.

The Qwen3-4B-Instruct results (Table~\ref{tab:ruler_qwen}) show that Pulsar remains close to Dense Attention without architecture-specific tuning, supporting cross-architecture applicability.

\begin{table}[t]
  \caption{RULER average accuracy (\%) on Qwen3-4B-Instruct.
    \textbf{Bold} = best per context length.}
  \label{tab:ruler_qwen}
  \begin{center}
    \begin{small}
      \begin{tabular}{lcccc}
        \toprule
        \textbf{Method} &
        \textbf{16K} & \textbf{32K} & \textbf{64K} &
        \textbf{128K} \\
        \midrule
        Full Attention
          & \textbf{92.4} & 89.3 & 88.6 & \textbf{86.5} \\
        \textbf{Pulsar Attention (Ours)}
          & 91.6 & \textbf{90.4} & \textbf{88.8} & 85.6 \\
        \bottomrule
      \end{tabular}
    \end{small}
  \end{center}
\end{table}

\subsection{Results on BABILong}
\label{sec:babilong}

Table~\ref{tab:babilong-16k} reports accuracy on three BABILong tasks (qa1, qa3, qa5) at 16K context length using $B = 4$ blocks of 4K tokens each. Pulsar Attention with Max-IDF scoring ($\sigma = 512$) achieves 37\% on qa1, 44\% on qa3, and 63\% on qa5. On the multi-hop tasks (qa3 and qa5), Pulsar outperforms Star Attention (33\% and 58\% respectively) by +11\% and +5\%. The advantage on qa3 is particularly notable: this task requires chaining three supporting facts that may reside in different blocks, and Max-IDF's ability to select chunks containing rare entity tokens provides cross-block context that Star's position-fixed anchor cannot.

\subsection{RULER Task-Category Analysis}
\label{sec:taskwise}

Table~\ref{tab:ruler_taskwise} reports per-task-category accuracy at 128K tokens on RULER, where the differences between methods are most pronounced.

\begin{table}[t]
  \caption{Per-task accuracy (\%) at 128K on Llama-3.1-8B-Instruct ($B=8$, block size 16K, summary budget 12.5\%). NIAH MultiKey and NIAH Single scores are averaged across their three sub-tasks; QA averages SQuAD and HotpotQA.}
  \label{tab:ruler_taskwise}
  \begin{center}
    \begin{small}
      \begin{tabular}{lccc}
        \toprule
        \textbf{Task} & \textbf{Dense} &
        \textbf{Star} & \textbf{Pulsar} \\
        \midrule
        NIAH Single (avg)    & 99.7 & 99.9 & \textbf{100.0} \\
        NIAH MultiKey (avg)  & \textbf{83.9} & 80.1 & 78.7 \\
        NIAH MultiValue      & \textbf{91.6} & 82.7 & 63.5 \\
        NIAH MultiQuery      & \textbf{97.8} & 96.6 & 97.0 \\
        Variable Tracking    & 61.8 & \textbf{62.8} & 61.0 \\
        Common Words         & 0.04 & 0.04 & \textbf{81.2} \\
        Frequent Words       & 72.3 & 75.9 & \textbf{80.7} \\
        QA (avg)             & \textbf{58.8} & 54.8 & 54.5 \\
        \midrule
        \textbf{Overall Avg} & 76.3 & 74.4 & \textbf{79.1} \\
        \bottomrule
      \end{tabular}
    \end{small}
  \end{center}
  \vskip -0.1in
\end{table}

\textbf{Retrieval (NIAH).} Single-needle retrieval is effectively saturated (~100\%) across all methods. On MultiKey and MultiQuery, Pulsar maintains near-parity with Dense, as Max-IDF reliably promotes chunks containing rare UUID or keyword needles. The notable exception is NIAH MultiValue (63.5\% vs. Dense 91.6\%): this task requires retrieving multiple values associated with a single key, where the values may reside in different blocks. Max-IDF selects the chunk containing the rare key identifier but may miss value chunks whose tokens are individually less distinctive.

\begin{figure}[t]
  \centering
  \includegraphics[width=\linewidth, height=5.07cm]{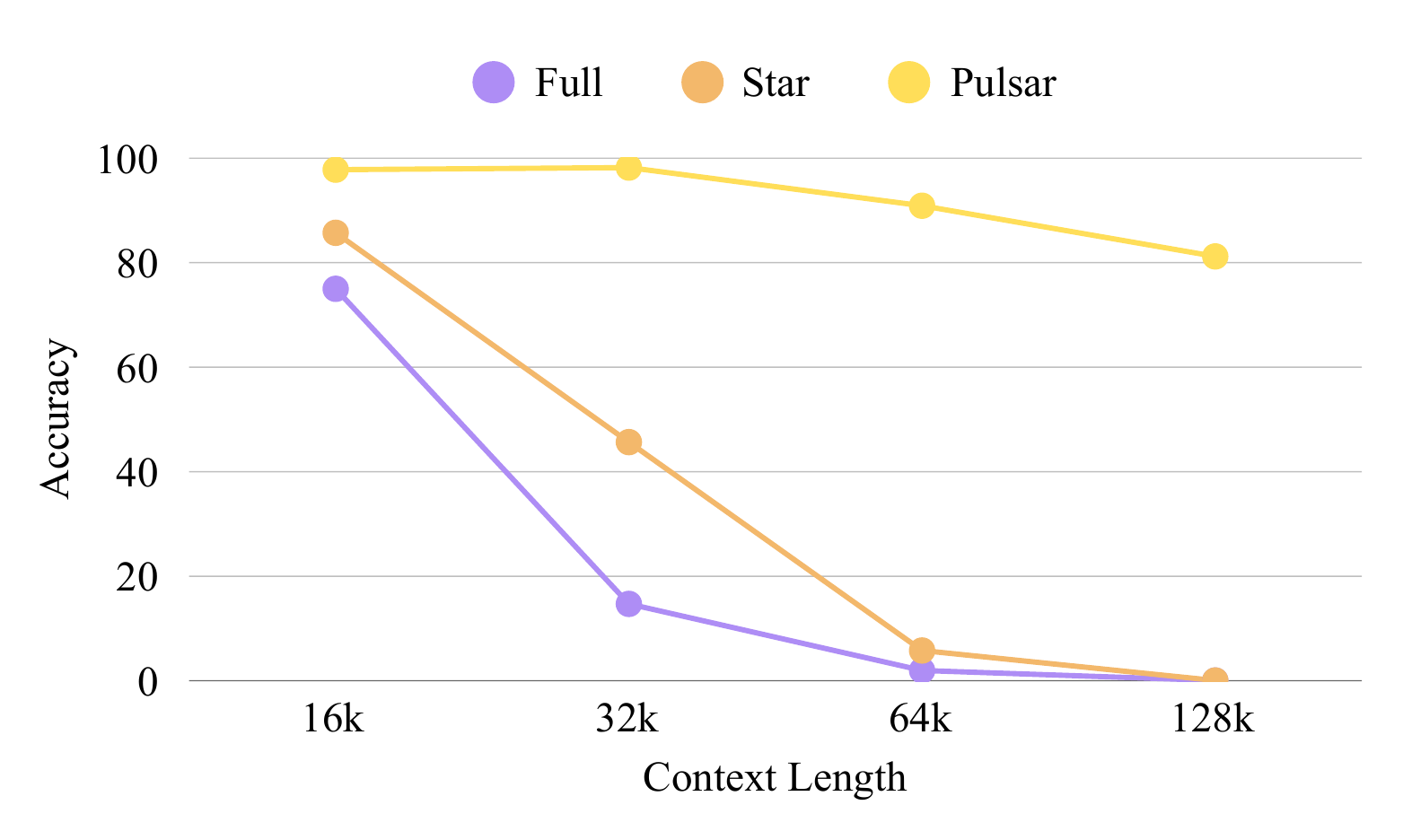}
  \caption{Accuracy on the Common Words Extraction task (RULER) across context lengths using Llama-3.1-8B-Instruct. Full Attention and Star Attention both collapse to near-zero accuracy at 128K, while Pulsar Attention maintains 81.2\% by propagating IDF-scored summaries that preserve cross-block frequency.}
  \label{fig:cwe}
\end{figure}

\begin{table*}[!t]
\caption{\textsc{BABILong}-16K accuracy (\%) across three summary budgets($\sigma$). \textbf{Bold}: best among the four IDF/entropy-based heuristics (TF-IDF, BM25, Entropy, Max-IDF) per budget–task group. Even-Spaced and Star Attention are shown as non-statistical baselines; Star uses a fixed size anchor block ($B=4$, block size 4K).}
\label{tab:babilong-16k}
\centering\small
\setlength{\tabcolsep}{5pt}
\begin{tabular}{l|rrr|rrr|rrr}
\toprule
& \multicolumn{3}{c|}{$\sigma = 128$}
& \multicolumn{3}{c|}{$\sigma = 512$}
& \multicolumn{3}{c}{$\sigma = 1{,}024$} \\
\cmidrule(lr){2-4}\cmidrule(lr){5-7}\cmidrule(lr){8-10}
Heuristic & qa1 & qa3 & qa5
          & qa1 & qa3 & qa5
          & qa1 & qa3 & qa5 \\
\midrule
TF-IDF
  & \textbf{40} & 33 & 61
  & 37          & 30 & 57
  & 33          & 30 & \textbf{60} \\
BM25
  & 38 & 34 & 61
  & \textbf{39} & 32 & 57
  & 33 & \textbf{35} & \textbf{60} \\
Entropy
  & 36 & \textbf{39} & 65
  & 38 & 39 & 62
  & 35 & 27 & \textbf{60} \\
Max-IDF \textit{(ours)}
  & 39 & 36 & \textbf{66}
  & 37 & \textbf{44} & \textbf{63}
  & \textbf{36} & 34 & 57 \\
\midrule
Even-Spaced & 49 & 36 & 57 & 44 & 26 & 55 & 46 & 25 & 52 \\
Star Attn   & 35 & 33 & 58 & 35 & 33 & 58 & 35 & 33 & 58 \\
\bottomrule
\end{tabular}
\end{table*}

\textbf{Aggregation.} Common Words Extraction reaches 81.2\% for Pulsar at 128K, compared with 0.04\% for Dense Attention and Star Attention; Frequent Words also improves (80.7\% vs. 72.3\% Dense and 75.9\% Star). This result is consistent with the hypothesis that causal summaries make repeated cross-block evidence easier to aggregate at long context lengths. The strongest gains over Dense Attention are therefore concentrated in frequency-aggregation tasks rather than being uniform across all task categories.

\textbf{Multi-hop reasoning.} Variable Tracking and QA show broad parity across all three methods (within 4\%), consistent with the expectation that multi-hop inference benefits from cross-block context but requires denser inter-block communication than the current summary budget provides.

\subsection{InfiniteBench Evaluation}
\label{sec:infinitebench}

Table~\ref{tab:infinitebench} evaluates 128K natural workloads: En.Sum is long-document summarization, En.MC is book-level multiple-choice comprehension, and Code.Debug is repository-level code debugging. Pulsar remains comparable to Dense Attention on long-document summarization, slightly improves on book-level comprehension, and matches Star Attention on repository-level code debugging, extending the evaluation beyond controlled synthetic workloads.

\begin{table}[t]
\caption{InfiniteBench accuracy (\%) at 128K.}
\label{tab:infinitebench}
\centering\small
\begin{tabular}{lrrr}
\toprule
Task & Dense & Star & Pulsar \\
\midrule
En.Sum & 31.91 & 31.85 & 31.76 \\
En.MC & 69.43 & 69.00 & 69.91 \\
Code.Debug & 16.75 & 24.37 & 24.43 \\
\bottomrule
\end{tabular}
\end{table}

\section{Ablations}
\label{sec:ablations}

\subsection{Scoring Heuristic Comparison}
\label{sec:ablation-heuristic}

Table~\ref{tab:babilong-16k} compares Max-IDF against three alternative scoring heuristics (TF-IDF, BM25, Entropy) and two non-statistical baselines (Even-Spaced sampling and Star Attention's fixed anchor) on BABILong-16K at three summary budgets ($\sigma \in \{128, 512, 1024\}$). Extended results at 32K and 64K are reported in Appendix~\ref{app:babilong-extended}.

At the $\sigma = 512$ budget, \textbf{Max-IDF} gives the strongest \texttt{qa3} and \texttt{qa5} performance among the evaluated heuristics: \texttt{qa3} rises to 44\% (+11\% over Star Attention) and \texttt{qa5} reaches 63\% (+5\% over Star Attention).

\paragraph{Why Max-IDF is effective at moderate budgets.}
Max-IDF is not uniformly optimal across all tasks and budgets: Even-Spaced performs better on \texttt{qa1}, where broad positional coverage is useful. At the selected 12.5\% budget, however, Max-IDF gives the strongest \texttt{qa3} and \texttt{qa5} performance among all evaluated heuristics, supporting its use for sparse multi-hop evidence selection.

\begin{table*}[!t]
\caption{Phase 1 per-layer computational cost for the critical (last) block on Llama-3.1-8B-Instruct ($B= 4$, $\sigma=512$, $s_{\text{sink}}=64$, bf16). All three columns: critical-path sequence length, peak activation memory, and FLOPs are reported per layer for the worst-case host.}
\label{tab:efficiency}
\centering\small
\resizebox{\textwidth}{!}{%
\setlength{\tabcolsep}{5.5pt}
\begin{tabular}{l|rrr|rrr|rrr}
\toprule
& \multicolumn{3}{c|}{Critical-path length $n$}
& \multicolumn{3}{c|}{Peak act.\ mem.\ / layer}
& \multicolumn{3}{c}{Phase~1 FLOPs / layer} \\
\cmidrule(lr){2-4}\cmidrule(lr){5-7}\cmidrule(lr){8-10}
Method   & 16K & 32K & 64K
         & 16K & 32K & 64K
         & 16K & 32K & 64K \\
\midrule
Dense (serial)
  & 16,384 & 32,768 & 65,536
  & 320\,MB & 640\,MB & 1.25\,GB
  & 2,749\,G & 10,995\,G & 43,981\,G \\
Star Attn
  &  8,192 & 16,384 & 32,768
  & 160\,MB & 320\,MB &  640\,MB
  &   687\,G &  2,749\,G & 10,995\,G \\
Pulsar \textit{(ours)}
  &  5,696 &  9,792 & 17,984
  & 117\,MB & 201\,MB &  368\,MB
  &   332\,G &    982\,G &  3,312\,G \\
\midrule
\multicolumn{10}{l}{\textit{Reduction vs.\ Dense}} \\[1pt]
Star Attn
  & $2.0\times$ & $2.0\times$ & $2.0\times$
  & $2.0\times$ & $2.0\times$ & $2.0\times$
  & $4.0\times$ & $4.0\times$ & $4.0\times$ \\
Pulsar \textit{(ours)}
  & $2.9\times$ & $3.4\times$ & $3.6\times$
  & $2.9\times$ & $3.4\times$ & $3.6\times$
  & $8.3\times$ & $11.2\times$ & $13.3\times$ \\
\midrule
\multicolumn{7}{l|}{\textit{Pulsar reduction vs.\ Star}}
  & $2.1\times$ & $2.8\times$ & $3.3\times$ \\
\bottomrule
\end{tabular}%
}
\end{table*}

\paragraph{Budget sensitivity and selection.} Max-IDF's advantage is not monotonic across all budgets. At $\sigma = 1024$, \texttt{qa3} drops from 44\% to 34\% and \texttt{qa5} from 63\% to 57\%; at $\sigma = 128$, Entropy slightly leads on \texttt{qa5} (66\% vs. Max-IDF's 65\%). We therefore fix $\sigma=12.5\%$ of block size (512 tokens for a 4K block), balancing selective multi-hop evidence coverage and Phase~1 overhead.

\subsection{Computational Efficiency}
\label{sec:ablation-efficiency}

For the evaluated configurations, Phase~1 cost depends jointly on summary size, logical block count $B$, and block-to-host mapping $\pi$; $B$ and the physical host count $H$ need not be equal. Table~\ref{tab:efficiency} reports the $B=4$ analytical comparison, while Appendix~\ref{app:scalability} gives the general host-level critical-path formulation and the $B=8,H=4$ case. After sink and summary KV entries are discarded, Star Attention and Pulsar have the same retained KV-cache footprint.

\paragraph{Measured wall-clock latency and GPU memory.}
Table~\ref{tab:wallclock} reports end-to-end wall-clock speedup and Table~\ref{tab:memory} reports total GPU memory reduction over Dense, measured on Llama-3.1-8B-Instruct at 16K--128K.

\begin{table}[!t]
\caption{Wall-clock speedup over Dense on Llama-3.1-8B-Instruct ($B=4$ for 16K–64K, $B=8$ for 128K). Latency is the mean end-to-end time per sample averaged over all 13 RULER tasks. \textbf{Bold}: best per context length.}
\label{tab:wallclock}
\centering\small
\setlength{\tabcolsep}{6pt}
\begin{tabular}{lcccc}
\toprule
\textbf{Method} & \textbf{16K} & \textbf{32K} & \textbf{64K} & \textbf{128K} \\
\midrule
Dense  & --- & --- & --- & --- \\
Star   & $1.72\times$ & $2.24\times$ & $2.92\times$ & $3.81\times$ \\
Pulsar & $\mathbf{2.00\times}$ & $\mathbf{2.92\times}$ & $\mathbf{4.27\times}$ & $\mathbf{6.24\times}$ \\
\bottomrule
\end{tabular}
\end{table}

\begin{table}[!t]
\caption{Total GPU memory reduction over Dense on Llama-3.1-8B-Instruct ($B=4$ for 16K–64K, $B=8$ for 128K), summed across all GPUs (fleet cost). Star Attention is omitted because it retains identical measured memory to Pulsar after Phase 1 KV discard.}
\label{tab:memory}
\centering\small
\setlength{\tabcolsep}{6pt}
\begin{tabular}{lcccc}
\toprule
\textbf{Method} & \textbf{16K} & \textbf{32K} & \textbf{64K} & \textbf{128K} \\
\midrule
Dense        & --- & --- & --- & --- \\
Pulsar  & $2.2\times$ & $3.0\times$ & $4.0\times$ & $5.1\times$ \\
\bottomrule
\end{tabular}
\end{table}

Pulsar's speedup over Dense grows from 2$\times$ at 16K to 6.24$\times$ at 128K, widening with context, while Star grows more slowly (1.72$\times$ to 3.81$\times$). This matches the analytical FLOPs estimates: Pulsar's fixed 1,600-token prefix is a diminishing fraction of block size as $L$ grows (Table~\ref{tab:efficiency}). Total fleet memory drops by 2.2$\times$ at 16K, rising monotonically to 5.1$\times$ at 128K; Star and Pulsar share identical measured peaks because both discard Phase~1 activations and retain only the local block KV cache. IDF table construction adds no measurable overhead, requiring only $O(L)$ integer operations on the CPU prior to any GPU computation.

\section{Related Work}
\label{sec:related}

\paragraph{Distributed long-context inference.}
Sequence parallelism \citep{korthikanti2023reducing} shards the sequence dimension with an all-reduce per layer. Ring Attention \citep{liu2023ring} removes this synchronization by overlapping communication with computation in a ring topology, but still requires cross-host transfers at every layer. Star Attention \citep{acharya2024star} eliminates Phase 1 communication entirely by prepending a static first-block anchor to each host. A concurrent line of work, APB \citep{huang2025apb}, also targets this distributed blockwise regime, but replaces the anchor with compressed ``passing blocks'' selected by trained retaining heads~\citep{huang2024locret} --- small MLPs that require a supervised training pass over curated long-context data (e.g., LongAlign) for every backbone before deployment. This ties APB's context-selection mechanism to the availability of suitable training data and a per-model training step, limiting its use as a drop-in replacement across new architectures. Pulsar instead pursues a training-free design: Max-IDF summaries are computed from corpus-level token statistics in a single CPU pass, with no labeled data, no proxy model, and no per-backbone tuning, making it immediately portable to new model families --- as we confirm with Qwen3-4B-Instruct (Section~\ref{sec:experiments}) without any architecture-specific adaptation.

\paragraph{Sparse and efficient attention.}
Longformer \citep{beltagy2020longformer} combines sliding-window attention with global tokens at $O(n)$ cost but uses fixed structural patterns. MInference \citep{jiang2024minference} dynamically selects sparse attention patterns per head during pre-filling, achieving strong results up to 64K but degrading at 128K in our experiments (Table \ref{tab:ruler_main}). FlexPrefill \citep{lai2025flexprefill} extends dynamic sparsity with query-dependent budget allocation; its gains are orthogonal to our distributed encoding phase. StreamingLLM \citep{xiao2023efficient} preserves attention-sink tokens for streaming generation; we adopt its sink mechanism as one component of our prefix but pair it with content-aware summaries rather than a recency-based sliding window. FlashAttention \citep{dao2022flashattention} provides the tiled block computation underlying all our Phase 1 forward passes.

\paragraph{KV cache compression and context selection.}
H2O \citep{zhang2023h2o}, SnapKV \citep{li2024snapkv}, and PyramidKV \citep{cai2024pyramidkv} evict low-importance KV entries during generation based on attention scores accumulated over prior tokens; these methods operate post-hoc on a fully computed KV cache and are orthogonal to Pulsar's Phase~1 selection. Because Pulsar's Phase~2 produces a standard distributed KV cache, these eviction strategies can be applied on top of Pulsar without modification, potentially compounding their savings with ours.

\paragraph{Prompt compression.}
LLMLingua \citep{jiang2023llmlingua} and LongLLMLingua \citep{jiang2024longllmlingua} use a small proxy LM to score and remove tokens from the prompt before the main forward pass, reducing input length at the cost of an additional neural inference step. RECOMP \citep{xu2023recomp} similarly condenses retrieved passages via extractive or abstractive summarization. Pulsar's Max-IDF scoring is intentionally non-neural: it relies on corpus-level token statistics computable in a single CPU pass with no proxy model, adding negligible overhead while drawing on classical IDF-based retrieval \citep{robertson2009probabilistic}. The key distinction is that Pulsar selects summaries \emph{per block} for a distributed encoding phase, whereas prompt-compression methods target single-GPU prefill reduction.
\section{Conclusion}
\label{sec:conclusion}

We introduced Pulsar Attention, a drop-in replacement for Star Attention's static anchor block that uses two lightweight, content-aware components: a 64-token attention-sink prefix and Max-IDF block summaries at 12.5\% of block size. On RULER with Llama-3.1-8B-Instruct, Pulsar outperforms both Dense and Star Attention at 32K--128K tokens, with gains of up to $+4.7\%$ over Dense and $+6.1\%$ over Star, while reducing Phase~1 per-GPU FLOPs by up to $3.3\times$ relative to Star. End-to-end wall-clock profiling and GPU memory measurements confirm that the analytical FLOPs advantage translates to practical speedups, with Pulsar achieving up to $6\times$ over Dense at 64K. Consistent gains on Qwen3-4B-Instruct confirm that the approach generalizes across model families. The retained KV cache is identical to Star Attention's, making Pulsar compatible with existing Phase~2 infrastructure and post-hoc KV compression methods. The main limitation is on tasks requiring retrieval of multiple values per key (NIAH MultiValue), where single-score chunk ranking can miss value-bearing chunks that lack globally rare tokens. Future work will address this through query-aware or multi-score selection strategies.
\section{Limitations}
\label{sec:limitations}

While Pulsar Attention significantly reduces distributed encoding overhead, the \textsc{Max-IDF} heuristic struggles to retrieve multiple recurring values associated with a single key (e.g., the NIAH MultiValue task). Because numerical values lack global token rarity, they receive low IDF scores and are often bypassed, necessitating future work on query-aware or multi-score selection strategies. 

Second, our evaluation relies on synthetic benchmarks (RULER) and embedded reasoning suites (\textsc{BABILong}). While rigorous, these do not fully capture the organic discourse structures and cross-document dependencies of real-world workloads, such as open-ended multi-document summarization or repository-level code reasoning. 

Finally, the summary budget ($\sigma = 12.5\%$) and chunk size ($m = 32$) are static global hyperparameters. Because attention dynamics vary systematically by model depth, an adaptive, layer-wise budget allocation could achieve a more optimal trade-off between Phase~1 FLOPs reduction and exact global attention recovery.
\section*{Acknowledgments}

We acknowledge the National Supercomputing Mission (NSM) for providing computing resources of `PARAM Ganga' at the Indian Institute of Technology Roorkee, which is implemented by C-DAC and supported by the Ministry of Electronics and Information Technology (MeitY) and Department of Science and Technology (DST), Government of India.

\bibliography{main}

\clearpage
\appendix

\begin{table*}[!t]
\caption{BABILong task descriptions.}
\label{tab:babilong-tasks}
\centering\footnotesize
\setlength{\tabcolsep}{6pt}
\renewcommand{\arraystretch}{1.1}
\begin{tabular}{lp{14cm}}
\toprule
Task & Description \\
\midrule
\texttt{qa1} & \textbf{Single Supporting Fact.} One sentence states where a
person currently is. The model must track the most recent location across
multiple movement facts and answer ``Where is $X$?'' \\
\texttt{qa3} & \textbf{Three Supporting Facts / Object Path.} Three facts
establish an object's path across locations. The model must answer ``Where was
the object \emph{before} location $Y$?'', requiring three-hop reasoning over
the chain. \\
\texttt{qa5} & \textbf{Three-Argument Relations.} Facts involve object-transfer
events (picked-up, gave, passed). The model answers queries such as ``Who gave
the apple?'' or ``What did $X$ give to $Y$?'', requiring entity-argument
co-reference across three-place predicates. \\
\bottomrule
\end{tabular}
\end{table*}

\begin{table*}[!t]
\caption{RULER task descriptions. Abbreviations: $k$ = number of unique
needle keys; $v$ = values per key; $r$ = number of queries.}
\label{tab:ruler-tasks}
\centering\footnotesize
\setlength{\tabcolsep}{6pt}
\renewcommand{\arraystretch}{1.05}
\begin{tabular}{llp{12cm}}
\toprule
Category & Task & Description \\
\midrule
\multirow{8}{*}{NIAH}
& Single-1 & Single word key $\to$ single number value hidden in synthetic noise text. \\
& Single-2 & Single word key $\to$ single number value hidden in Paul Graham essay excerpts. \\
& Single-3 & Single word key $\to$ UUID value hidden in book text. UUID tokens are maximally rare (high IDF). \\
& MultiKey-1 & 4 distinct word keys all map to the same number; the model must retrieve the value from any key. \\
& MultiKey-2 & An entire line of word keys maps to one value; model must match the full key phrase. \\
& MultiKey-3 & UUID keys $\to$ UUID values; both key and value are rare tokens. \\
& MultiValue & 1 key maps to 4 different numbers; the model must recall all of them. \\
& MultiQuery & 4 queries each asking for the value of a different key from the same set of needles. \\
\midrule
Tracking & Variable & A chain of variable assignments is hidden in noise (e.g., \texttt{x=3; y=x; z=y}).
The model finds all variables assigned a target value. \\
\midrule
\multirow{2}{*}{Words}
& CWE & A numbered list of words is presented; some appear more often.
The model identifies the 10 most common words. \\
& FWE & A ``coded'' text with high-frequency synthetic words.
The model identifies the 3 most frequent codes. \\
\midrule
\multirow{2}{*}{QA}
& QA1 & Reading comprehension over SQuAD~\citep{rajpurkar2016squad} passages concatenated into a long context. \\
& QA2 & Multi-hop reading comprehension over HotpotQA~\citep{yang2018hotpotqa} documents; requires reasoning across two documents. \\
\bottomrule
\end{tabular}
\end{table*}

\section{Dataset Details}
\label{app:tasks}

\subsection{BABILong}
\label{app:babilong}

BABILong~\citep{kuratov2024babilong} extends the bAbI reasoning suite by
embedding short synthetic reasoning chains inside long distractor text drawn
from PG-19 books. The reasoning chains are generated from symbolic rules; the
model must locate the relevant sentences among thousands of distractor tokens
and apply multi-step logic. We evaluate on two tasks;
Table~\ref{tab:babilong-tasks} summarises their requirements.

Context splits cover $L \in \{16\text{K},\,32\text{K},\,64\text{K},\,128\text{K}\}$ tokens;
each split contains 1,000 samples. We report results for \texttt{qa1},
\texttt{qa3}, and \texttt{qa5} as representative of increasing reasoning depth.

\subsection{RULER}
\label{app:ruler}

RULER~\citep{hsieh2024ruler} is a synthetic benchmark designed to test
long-context recall under controlled conditions. We use all 13 tasks across
four context lengths
($L \in \{16\text{K},\,32\text{K},\,64\text{K},\,128\text{K}\}$),
500 samples per task per length. Table~\ref{tab:ruler-tasks} describes each task.

\section{Chat Template and Prompt Format}
\label{app:template}

\subsection{Llama~3 Template}
\label{app:template-llama}

Llama experiments use Meta-Llama-3.1-8B-Instruct with the standard Llama~3
instruct chat template. The full template skeleton is:

\begin{small}
\begin{verbatim}
<|begin_of_text|>
<|start_header_id|>system<|end_header_id|>

You are a helpful assistant.<|eot_id|>
<|start_header_id|>user<|end_header_id|>

{task_instruction}

{in_context_examples}

{post_prompt}

<context>
{long_context}
</context>

Question: {query}<|eot_id|>
<|start_header_id|>assistant<|end_header_id|>

\end{verbatim}
\end{small}

\paragraph{Stop words (Llama~3).}
Generation halts at any of: \texttt{<|end\_of\_text|>}, \texttt{<|eom\_id|>},
\texttt{<|eot\_id|>}, or after \texttt{max\_new\_tokens} (128 for BABILong;
task-dependent for RULER, see Table~\ref{tab:hyperparams}).

\paragraph{Tokenisation.}
Tokens are produced with \texttt{add\_special\_tokens=False}; the
\texttt{<|begin\_of\_text|>} BOS token is included as part of the template
string itself and must not be added a second time.

\subsection{Qwen3 Template}
\label{app:template-qwen}

Qwen3 experiments use Qwen3-4B-Instruct~\citep{qwen3} with the ChatML
instruct template. Thinking mode is disabled throughout all experiments.
The full template skeleton is:

\begin{small}
\begin{verbatim}
<|im_start|>system
You are a helpful assistant.<|im_end|>
<|im_start|>user

{task_instruction}

{in_context_examples}

{post_prompt}

<context>
{long_context}
</context>

Question: {query}<|im_end|>
<|im_start|>assistant

\end{verbatim}
\end{small}

\paragraph{Stop words (Qwen3).}
Generation halts at: \texttt{<|im\_end|>} or \texttt{<|endoftext|>}, or after
\texttt{max\_new\_tokens} (same per-task values as Table~\ref{tab:hyperparams}).

\paragraph{Tokenisation.}
As with Llama~3, tokens are produced with \texttt{add\_special\_tokens=False};
the system header is embedded directly in the template string.

\subsection{Context/Query Split}
\label{app:template-split}

Both models share the same two-string decomposition before tokenisation:

\begin{itemize}[leftmargin=1.5em,itemsep=2pt]
  \item \textbf{\texttt{prompt\_context}}: everything from the opening header
    token through the closing \texttt{</context>} tag, inclusive. This string
    is tokenised, split into blocks of size $B$, and processed by Phase~1.
  \item \textbf{\texttt{prompt\_query}}: the \texttt{Question:~\{query\}}
    fragment followed by the assistant header tokens. This string is tokenised
    separately and processed by Phase~2.
\end{itemize}
\section{Scoring Heuristic Details}
\label{app:heuristics}

All heuristics operate on non-overlapping contiguous chunks of
\texttt{chunk\_size} tokens carved from each block. Chunks preserve local word
order; the top-\texttt{num\_chunks} chunks by score are selected and returned
\emph{in positional order} to maintain natural reading order for the
Transformer.

\subsection{Corpus-Level IDF Table}
\label{app:idf}

Before any scoring, a single pass over all blocks constructs a
document-frequency table. Let $N$ be the total number of blocks and
$\mathrm{df}(t)$ the number of blocks containing token $t$:
\begin{equation}
  \mathrm{IDF}(t) = \log\!\left(\frac{N}{\max(\mathrm{df}(t),\,1)}\right).
  \label{eq:idf}
\end{equation}
This is stored as a dense integer-indexed tensor \texttt{idf\_table} of shape
$(\texttt{vocab\_size},)$. All four IDF-based heuristics perform a single
gather from this table; no Python loops over the vocabulary are required.

\subsection{TF-IDF}
\label{app:tfidf}

\begin{equation}
  \text{score}_{\text{TF-IDF}}(C)
    = \frac{1}{|C|} \sum_{p \in C} \mathrm{IDF}(t_p).
  \label{eq:tfidf}
\end{equation}
This equals the classical mean TF-IDF over types, and permits a single
vectorized lookup:
\texttt{idf\_table[}\allowbreak\texttt{chunk\_tokens]}\allowbreak\texttt{.sum() / len(chunk)}.

\textbf{Strengths.} Balanced, general-purpose; promotes chunks with rare but
recurrent vocabulary. \textbf{Weakness.} A single rare entity among many common
tokens is diluted by the average.

\subsection{BM25}
\label{app:bm25}

\begin{equation}
\begin{aligned}
  \text{score}_{\text{BM25}}(C)
    = \sum_{t \in \text{types}(C)} \!\!
        & \mathrm{IDF}_{\text{BM25}}(t)\;\times \\[-2pt]
        & \frac{\mathrm{tf}(t)\,(k_1+1)}
               {\mathrm{tf}(t) + k_1 D(C)},
\end{aligned}
\label{eq:bm25}
\end{equation}
where $D(C) = 1 - b + b\,\tfrac{|C|}{\overline{l}}$, $k_1\!=\!1.2$,
$b\!=\!0.75$, $\overline{l}$ is the mean chunk length across the corpus, and
\begin{equation*}
  \mathrm{IDF}_{\text{BM25}}(t) = \log\!\left(\frac{N - \mathrm{df}(t) + 0.5}{\mathrm{df}(t) + 0.5} + 1\right).
\end{equation*}

\textbf{Strengths.} More robust than TF-IDF when chunk lengths vary.
\textbf{Weakness.} Slightly higher compute; benefits diminish when all chunks
are the same length.

\subsection{Entropy (Type-Token Ratio)}
\label{app:entropy}

\begin{equation}
  \text{score}_{\text{Entropy}}(C)
    = \frac{\bigl|\{t : t \in C\}\bigr|}{|C|}.
  \label{eq:entropy}
\end{equation}
Implemented as \texttt{(bincount(C) > 0).sum() / len(C)}.

\textbf{Strengths.} Corpus-independent; no IDF table required. A chunk of
padding tokens or repeated stop-words scores near 0; an entity-rich chunk
scores near 1. \textbf{Weakness.} Ignores global rarity; a chunk with 32
distinct common stop-words scores identically to one with 32 rare entities.

\subsection{Max-IDF}
\label{app:maxidf}

\begin{equation}
  \text{score}_{\text{Max-IDF}}(C)
    = \max_{p \in C}\, \mathrm{IDF}(t_p).
  \label{eq:maxidf}
\end{equation}
Implemented as
\texttt{idf\_table[}\allowbreak\texttt{chunk\_tokens]}\allowbreak\texttt{.max()}.

\textbf{Strengths.} Acts as a pure needle detector: if any token in the chunk
appears in only one block, the chunk is guaranteed to be selected. This aligns with what multi-hop tasks and NIAH require. \textbf{Weakness.} Ignores
overall chunk quality; one rare token among 31 uninformative tokens still wins.

\subsection{Even-Spaced (Baseline)}
\label{app:evenly-spaced}

Selects exactly $\texttt{num\_chunks} \times \texttt{chunk\_size}$ tokens at
uniform intervals across the block using \texttt{torch.linspace}:
\begin{equation}
\begin{aligned}
  \text{indices} &= \lfloor \texttt{linspace}(0,\, |B|-1,\, s_{\text{sink}}) \rfloor, \\
  \sigma &= \texttt{num\_chunks} \times \texttt{chunk\_size}.
\end{aligned}
\end{equation}
Original (sparse) position IDs are preserved.

\textbf{Strengths.} No corpus statistics required; deterministic. \textbf{Weakness.} Wastes budget on low-information spans as budget grows (confirmed in
ablations: \texttt{qa3} degrades from 36 to 25 as $\sigma$ increases from 128 to
1024).

\subsection{Heuristic Summary}
\label{app:heuristic-summary}

Table~\ref{tab:heuristic-summary} consolidates the five heuristics described
above, listing the signal each one uses, whether it requires corpus-level
statistics, the task profile it suits best, and its main weakness.

\begin{table*}[!t]
\caption{Summary of all scoring heuristics.}
\label{tab:heuristic-summary}
\centering\small
\setlength{\tabcolsep}{8pt}
\begin{tabular}{lllll}
\toprule
Heuristic & Signal & Corpus-dep.\ & Best for & Weakness \\
\midrule
TF-IDF   & Mean IDF          & Yes & General-purpose      & Single rare token diluted \\
BM25     & Len-norm.\ IDF    & Yes & Variable chunk sizes & Extra \texttt{bincount} cost \\
Entropy  & Type-token ratio  & No  & Small corpus / fast  & Blind to global rarity \\
Max-IDF  & Peak IDF          & Yes & Rare entity / NIAH   & Ignores chunk quality \\
Even-Sp. & Uniform coverage  & No  & Coverage guarantee   & Wastes budget on filler \\
\bottomrule
\end{tabular}
\end{table*}

\section{Hyperparameter Configuration}
\label{app:hyperparams}

Table~\ref{tab:hyperparams} lists all hyperparameters used in our main
experiments. All values are fixed across RULER and BABILong unless noted.

\begin{table*}[!t]
\caption{Full hyperparameter configuration.}
\label{tab:hyperparams}
\centering\small
\setlength{\tabcolsep}{8pt}
\begin{tabular}{lll}
\toprule
Parameter & Value & Notes \\
\midrule
\multicolumn{3}{l}{\textit{Model — Llama~3}} \\
Model & Meta-Llama-3.1-8B-Instruct & \\
Precision & bfloat16 & \\
Attention impl. & FlashAttention-2 & \texttt{attn\_impl=flash\_attention\_2} \\
Decoding & Greedy (argmax) & Temperature = 0 \\
\midrule
\multicolumn{3}{l}{\textit{Model — Qwen3}} \\
Model & Qwen3-4B-Instruct~\citep{qwen3} & \\
Precision & bfloat16 & \\
Attention impl. & FlashAttention-2 & GQA: 32 Q-heads / 8 KV-heads \\
Thinking mode & Disabled & \\
Decoding & Greedy (argmax) & Temperature = 0 \\
\midrule
\multicolumn{3}{l}{\textit{Distributed setup}} \\
Number of GPUs ($H$) & 4 & $B$ blocks split evenly between hosts \\
Block size & 25\% of context length & 12.5\% of length at 128K for memory efficiency \\
\midrule
\multicolumn{3}{l}{\textit{Pulsar Attention: summary}} \\
Summary method & Max-IDF & Eq.~\eqref{eq:maxidf} \\
Summary budget ($\sigma$) & 512 tokens / block (at 16K) & 12.5\% of block size \\
Chunk size & 32 tokens & 16 chunks selected per block \\
Sink size ($s_{\text{sink}}$) & 64 tokens & First 64 tokens of block~0 \\
KV discard & Enabled & Non-$B_i$ KV sliced after Phase~1 \\
Position mode & Sparse & Original global position IDs preserved \\
\midrule
\multicolumn{3}{l}{\textit{Generation}} \\
Max new tokens (BABILong) & 128 & \\
Max new tokens (RULER NIAH) & 128 & \\
Max new tokens (RULER VT) & 30 & \\
Max new tokens (RULER CWE) & 120 & \\
Max new tokens (RULER FWE) & 50 & \\
Max new tokens (RULER QA) & 32 & \\
\bottomrule
\end{tabular}
\end{table*}

\section{RoPE Position ID Assignment}
\label{app:rope}

When a block's Phase~1 input is assembled as
$[\text{SINK} \mid S_1,\ldots,S_{i-1} \mid B_i]$, the position IDs of the
inserted sink and summary tokens can be assigned in two ways.

\paragraph{Sparse (default).}
Every token retains its \emph{original global position} in the full context.
Summary tokens from block~$j$ keep the position IDs of their source locations.
The assembled sequence therefore contains \emph{gaps} where non-selected tokens
were dropped. This preserves the true inter-token distances that RoPE was
trained to encode: a summary token from block~0 is still seen as far from a
token in block~3, matching the model's pre-training geometry.

\paragraph{Contiguous.}
The assembled sequence is re-numbered $0, 1, \ldots, L_i - 1$ for each block
$i$'s Phase~1 forward pass (original Star Attention style). In this mode,
Phase~2 query position IDs are offset to start just past the longest assembled
Phase~1 length, ensuring RoPE deltas from the query to every block's KV remain
small and in-distribution.

Contiguous mode can occasionally help by ensuring that summary tokens and local
block tokens are seen at short relative distances, but it distorts the global
geometry. All experiments reported in this paper use \textbf{sparse} mode,
which aligns with the model's pre-training distribution.

\section{KV Cache Discard}
\label{app:kv-discard}

After each Phase~1 forward pass, the model's KV cache contains entries for
\emph{all} tokens in the assembled input
$[\text{SINK} \mid \Sigma_0,\ldots,\Sigma_{i-1} \mid B_i]$. The sink and summary KV
states are discarded immediately:
\begin{equation}
\begin{aligned}
  K_i^{\text{(layer)}} &\leftarrow K_i^{\text{(layer)}}[\,:,:,\texttt{is\_bi}\,], \\
  V_i^{\text{(layer)}} &\leftarrow V_i^{\text{(layer)}}[\,:,:,\texttt{is\_bi}\,],
\end{aligned}
\end{equation}
where \texttt{is\_bi}$[p] = \text{True}$ iff $p \geq \texttt{summary\_len}$.
Implemented via a boolean mask:

\begin{small}
\begin{verbatim}
is_bi = torch.zeros(total_len, dtype=torch.bool)
is_bi[-block_size:] = True
kv = [[x[0][:,:,is_bi], x[1][:,:,is_bi]] for x in kv]
\end{verbatim}
\end{small}

Discarding serves two purposes: \textbf{(1)}~\emph{No duplicate tokens in
Phase~2}: summary token $S_j$ was prepended to blocks $j{+}1,\ldots,B{-}1$;
if retained, each KV cache would contain a copy of $\sigma_j$'s key/value vectors,
corrupting the softmax normalisation during Phase~2's \texttt{all\_gather}.
\textbf{(2)}~\emph{Exact final cache size}: after discarding, each GPU holds
exactly \texttt{block\_size} KV entries, so the total KV cache across $B$ GPUs
equals the original context length $L$.

\section{Extended BABILong Heuristic Results}
\label{app:babilong-extended}

Tables~\ref{tab:babilong-32k} and~\ref{tab:babilong-64k} extend the 16K
heuristic comparison (Table~\ref{tab:babilong-16k} in the main paper) to
32K and 64K context lengths, at both the 3.125\% ($\sigma\!=\!128$) and 12.5\%
($\sigma\!=\!512$) summary budgets ($B\!=\!4$, block sizes 8K and 16K respectively).
At the 12.5\% budget ($s\!=\!512$), Max-IDF is best on multi-hop tasks
(\texttt{qa3}, \texttt{qa5}) across both 32K and 64K.
At the smaller 3.125\% budget ($\sigma\!=\!128$), averaging heuristics
(TF-IDF, BM25) lead on \texttt{qa3}: at this budget the fixed chunk
count is small enough that selecting diverse representative chunks
outweighs pure needle-detection.
Max-IDF dominates \texttt{qa5} at 32K--3.125\% but not at 64K--3.125\%,
where \texttt{qa5} scores are uniformly lower and BM25/Entropy tie for best.
Star Attention was not evaluated at 64K and its cells are left blank.

\begin{table*}[!t]
\caption{%
  \textsc{BABILong}-32K accuracy (\%) across two summary budgets
  ($B\!=\!4$, block size 8\,K).
  \textbf{Bold}: best among the four statistical heuristics per group.
  Star Attention uses a fixed anchor block independent of summary budget.
}
\label{tab:babilong-32k}
\centering\small
\setlength{\tabcolsep}{6pt}
\begin{tabular}{l|rrr|rrr}
\toprule
& \multicolumn{3}{c|}{$\sigma = 128$ (3.125\%)}
& \multicolumn{3}{c}{$\sigma = 512$ (12.5\%)} \\
\cmidrule(lr){2-4}\cmidrule(lr){5-7}
Heuristic & qa1 & qa3 & qa5
          & qa1 & qa3 & qa5 \\
\midrule
TF-IDF   & \textbf{34} & \textbf{17} & \textbf{48} & \textbf{32} & 15 & 48 \\
BM25     & 28 & \textbf{17} & 45 & 29 & 17 & 46 \\
Entropy  & 21 & 15 & 42 & 30 & 15 & 45 \\
Max-IDF \textit{(ours)} & 28 & 10 & 43 & 31 & \textbf{19} & \textbf{49} \\
\midrule
Even-Spaced & 33 & 17 & 55 & 30 & 16 & 48 \\
Star Attn   & 28 & 13 & 41 & 28 & 13 & 41 \\
\bottomrule
\end{tabular}
\end{table*}

\begin{table*}[!t]
\caption{%
  \textsc{BABILong}-64K accuracy (\%) across two summary budgets
  ($B\!=\!4$, block size 16\,K).
  \textbf{Bold}: best among the four statistical heuristics per group.
}
\label{tab:babilong-64k}
\centering\small
\setlength{\tabcolsep}{6pt}
\begin{tabular}{l|rrr|rrr}
\toprule
& \multicolumn{3}{c|}{$\sigma = 128$ (3.125\%)}
& \multicolumn{3}{c}{$\sigma = 512$ (12.5\%)} \\
\cmidrule(lr){2-4}\cmidrule(lr){5-7}
Heuristic & qa1 & qa3 & qa5
          & qa1 & qa3 & qa5 \\
\midrule
TF-IDF   & \textbf{28} &  9 & 35 & 17 &  5 & 31 \\
BM25     & 26 & \textbf{10} & \textbf{38} & 15 &  4 & 30 \\
Entropy  & 22 &  8 & \textbf{38} & 20 &  6 & 36 \\
Max-IDF \textit{(ours)} & 22 &  7 & 33 & \textbf{28} & \textbf{11} & \textbf{40} \\
\midrule
Even-Spaced & 28 &  2 & 56 & 24 &  5 & 34 \\
\bottomrule
\end{tabular}
\end{table*}

\section{Phase~2 Distributed Softmax Merge}
\label{app:phase2}

Phase~2 runs local attention over retained shards on all $H$ physical hosts. Each
host computes a local attention output $\mathbf{o}_h$ and log-sum-exp $\ell_h$
over its retained KV entries. After \texttt{dist.all\_gather}, the outputs are merged
with the numerically stable online-softmax
formula~\citep{milakov2018online}:
\begin{align}
  \ell' &= \ell + \log\!\left(1 + e^{\ell_h - \ell}\right),
  \label{eq:lse-merge} \\
  \mathbf{o}' &= e^{\ell - \ell'}\,\mathbf{o}
                + e^{\ell_h - \ell'}\,\mathbf{o}_h,
  \label{eq:out-merge}
\end{align}
applied iteratively over all gathered outputs. Equations
\eqref{eq:lse-merge}--\eqref{eq:out-merge} are implemented as a \texttt{torch.jit.script} kernel for
efficiency. The result is mathematically equivalent to running full attention
over the concatenated retained KV shards, at the cost of one
\texttt{all\_gather} communication round per generation step.

\section{Efficiency Measurement}
\label{app:efficiency}

All timing and peak-memory experiments are conducted on a single node equipped
with 4 $\times$ A100-80\,GB GPUs interconnected via NVLink, using bfloat16
precision and FlashAttention-2.

\paragraph{Wall-clock speedup.}
For each configuration (dense full attention, Star Attention, Pulsar Attention),
$N$ generation samples are processed sequentially. Per-sample time is measured
with a high-resolution monotonic clock started immediately before the forward
pass and stopped after a cross-rank synchronisation barrier, ensuring that the
recorded time reflects the slowest GPU (i.e.\ the true end-to-end latency).
Speedup is reported as the ratio of the mean per-sample wall-clock time under
dense attention to the mean per-sample time under the respective method.
Dataset loading, tokenisation, and IDF-table construction are excluded from
all timings.

\paragraph{Peak VRAM.}
Per-device peak memory statistics are reset before each sample. After
generation completes, peak GPU memory is read from the device allocator,
converted to gigabytes, and recorded per rank. We report two aggregates:

\begin{itemize}[leftmargin=1.5em,itemsep=2pt]
  \item \textbf{Per-GPU peak}: maximum recorded peak across all $H$ ranks,
    reflecting the most memory-pressured device.
  \item \textbf{Total peak}: sum of per-rank peaks across all $H$ GPUs,
    reflecting the aggregate KV-cache footprint of the full context.
\end{itemize}

For dense attention (single-process, multi-GPU via \texttt{device\_map=auto}),
the total peak is computed as the sum over all visible CUDA devices on the node,
ensuring comparability with the distributed total for the other methods.

\section{Block-to-Host Scalability Analysis}
\label{app:scalability}

The main paper uses $B$ for logical blocks, $H$ for physical hosts, and $\pi$
for the block-to-host mapping. For logical block $i$, its Phase~1 input length is
\begin{equation}
n_i = \frac{L}{B} + s_{\mathrm{sink}} + i\sigma.
\end{equation}
If $F(\cdot)$ denotes the per-block Phase~1 cost, host-level cost and the
critical path are
\begin{equation}
C_h^{\mathrm{Pulsar}}=\sum_{i:\pi(i)=h}F(n_i), \qquad
C_{\mathrm{crit}}^{\mathrm{Pulsar}}=\max_h C_h^{\mathrm{Pulsar}}.
\end{equation}
Increasing $H$ at fixed $B$ can reduce the number of blocks processed by a
host. Increasing $B$ at fixed $H$ creates more causal summaries and may increase
the host-level critical path. Thus cost depends jointly on summary size, logical
block count, and $\pi$; the $B=4$ derivation in the main text is a special case.
At 128K we use $B=8$, $H=4$, with two logical blocks processed sequentially on
each host. Larger host-count and topology sweeps, including crossovers with Star
Attention, remain future work.

\section{Additional Limitations and Protocol Details}
\label{app:limitations}

The frequency-aggregation mechanism is hypothesized rather than causally
established. Max-IDF is not uniformly best, and query-independent selection may
miss query-specific evidence. We fix the summary budget and chunk size, evaluate
natural workloads only on the three InfiniteBench tasks in the main paper, and
do not evaluate larger host-count or topology sweeps. En.Sum, En.MC, and
Code.Debug respectively measure long-document summarization, book-level
multiple-choice comprehension, and repository-level code debugging.

\end{document}